\def\year{2022}\relax
\documentclass[letterpaper]{article} 
\usepackage{aaai22}  
\usepackage{times}  
\usepackage{helvet}  
\usepackage{courier}  
\usepackage[hyphens]{url}  
\usepackage{graphicx} 
\usepackage{natbib}  
\usepackage{caption} 
\DeclareCaptionStyle{ruled}{labelfont=normalfont,labelsep=colon,strut=off} 
\usepackage{multirow}
\usepackage{booktabs}
\usepackage{amsmath}
\usepackage{bbding}
\usepackage{makecell}
\usepackage{amsfonts}
\usepackage{comment}
\usepackage{xcolor}
\usepackage{epstopdf}

\usepackage{algorithm}
\usepackage{algorithmic}

\usepackage{newfloat}
\usepackage{listings}
\floatstyle{ruled}
\newfloat{listing}{tb}{lst}{}
\floatname{listing}{Listing}
\title{U$^2$-Former: A Nested U-shaped Transformer for Image Restoration}
\author{
    Haobo Ji,
    Xin Feng,
    Wenjie Pei,
    Jinxing Li, 
    Guangming Lu, \emph{Member, IEEE}
}
\affiliations{

}

\usepackage{bibentry}

\begin{document}

\maketitle

\begin{abstract}
While Transformer has achieved remarkable performance in various high-level vision tasks, it is still challenging to exploit the full potential of Transformer in image restoration. The crux lies in the limited depth of applying Transformer in the typical encoder-decoder framework for image restoration, resulting from heavy self-attention computation load and inefficient communications across different depth (scales) of layers. In this paper, we present a deep and effective Transformer-based network for image restoration, termed as U$^2$-Former, which is able to employ Transformer as the core operation to perform image restoration in a deep encoding and decoding space. Specifically, it leverages the nested U-shaped structure to facilitate the interactions across different layers with different scales of feature maps. Furthermore, we optimize the computational efficiency for the basic Transformer block by introducing a feature-filtering mechanism to compress the token representation. Apart from the typical supervision ways for image restoration, our U$^2$-Former also performs contrastive learning in multiple aspects to further decouple the noise component from the background image. Extensive experiments on various image restoration tasks, including reflection removal, rain streak removal and dehazing respectively, demonstrate the effectiveness of the proposed U$^2$-Former.

\end{abstract}

\section{Introduction}

Image restoration is an important yet challenging research problem involving many tasks in computer vision, such as image reflection removal, image deraining and image dehazing. To efficiently reconstruct the image without corruption, accurate perception on  diverse noise patterns plays a key role. Most existing state-of-the-art methods~\cite{ronneberger2015u,fan2017generic,9424463} for image restoration are modeled based on the CNN structure due to its excellent performance of feature learning. Stemming from the inherent nature of the convolutional operation, an potential limitation for these methods is that the noise patterns are recognized only relying on the features learned in the local view of the image. Nevertheless, it is crucial to obtain a global perception of the whole image when performing image restoration. 

Unlike CNN that focuses on learning local shift-invariant features and expands the receptive field progressively by stacking convolutional layers, Transformer~\cite{vaswani2017attention} extracts features in a global view by its core operation, namely self-attention. As a result, a prominent benefit of Transformer, compared to CNN, is that each hidden unit in every feature learning layer involves the global context information of the input. Such characteristic makes Transformer particularly favorable to image-to-image mapping tasks, since the spatial coherence and the synthesizing patterns (e.g., the noise and background patterns in image restoration) tend to be learned easier in the global view. 

While Transformer has achieved remarkable progress in various high-level computer vision tasks~\cite{liu2021swin,carion2020end}, Transformer has not been extensively studied to exploit its full potential in image restoration. Recently, Uformer~\cite{wang2021uformer} applies Transformer to image restoration by embedding the self-attention block of Transformer into a U-shaped structure, thus the background image can be reconstructed by decoding features from different scales of feature maps, as performed by U-net~\cite{ronneberger2015u}. Whilst Uformer has shown promising performance in image restoration, a potential drawback is that the self-attention layer can only be applied in a limited depth in the U-shaped structure, which inevitably restricts the capability of noise pattern recognition and thus adversely affects the performance of image restoration. This is mainly resulted from two factors: 1) deeper U-shaped structure makes the communication across different depth (scales) of layers harder and thus hampers the model optimization via gradient back-propagation; 2) the heavy computation load of self-attention operation limits the depth of applying Transformer in the typical encoder-decoder framework for image restoration.

To address above limitation of Uformer, in this paper we propose the U$^2$-Former for image restoration which is also designed based on Transformer. Compared to Uformer and other existing methods for image restoration, our model benefits from three following advantages:
\begin{itemize}
\item An effective architecture is designed for image restoration, which enables our U$^2$-Former to employ Transformer as the core operation to construct both deep encoding and decoding space for image restoration. Specifically, the proposed U$^2$-Former leverages the nested U-shaped structure to facilitate the interactions across different layers with different scales of feature maps. Two nested U-shaped structures are adopted: the inner U-shaped Transformer block is built based on basic self-attention block and is responsible for aggregating features from different scales of feature maps. The outer U-shaped Encoder-Decoder framework utilizes the inner U-shaped Transformer block to construct the deep encoding and decoding space for learning noise patterns and separating noise from the background image.
To optimize the computational efficiency for the basic Transformer block, we propose a feature-filtering mechanism to compress the token representation by filtering out low-quality features. Benefiting from these two techniques, our U$^2$-Former is able to stack the Transformer block deeply to construct sufficiently deep feature space for separating noise and background.
\item Our U$^2$-Former performs multi-view contrastive learning to further decouple the noise component from the background image. In particular, contrastive learning is conducted in three aspects (views): 1) two patches from the same restored background image are viewed as positive pairs to ensure the restoring consistency between different regions in the restored image; 2) pairing the restored background image and the corresponding groundtruth background (with same image content) in patch level to be positive to guide the model to restore the clean background image; 3) comparing the restored background image to a random groundtruth background image (with different image content) in patch level as positive pairs to encourage the model to learn noise-sensitive features that is irrelevant to image content. Note that we construct negative pairs by comparing the restored background image and the restored noise image in all three cases.
\item We conduct extensive experiments on three image restoration tasks, including reflection removal, rain streak removal and image dehazing, to evaluate our U$^2$-Former, which show that our model consistently outperforms state-of-the-art methods for all three tasks.c
\end{itemize}

\section{Related Work}
\textbf{Image Restoration.} Stacking multiple convolutional layers is the well-known CNN-based strategy for image restoration tasks. For example, residual learning \cite{he2016deep} has been widely used for image reflection removal \cite{fan2017generic}, deraining \cite{yang2017deep}, and dehazing \cite{du2018recursive}. Similarly, extracting multi-scale information for capturing richer global context is also employed for image reflection removal \cite{wei2019single}. The encoder-decoder structure \cite{ronneberger2015u}, dense connections \cite{huang2017densely} and dilated convolution \cite{yu2015multi} are also very common in generic image restoration \cite{chen2019gated, feng2021contrastive}. Furthermore, attention mechanisms including spatial attention \cite{zhao2018psanet}, channel attention \cite{hu2018squeeze}, or both \cite{woo2018cbam}, perform a noticeable role in image restoration tasks \cite{li2020single, jiang2020multi, liu2019griddehazenet}, since the attention module enable the network to capture long-range global dependencies along spatial dimensions or channel dimensions.

\noindent\textbf{Vision Transformer.} Inspired by the applications of the Transformer in the natural language processing (NLP), numerous researchers have tried to introduce the Transformer in vision tasks. For image restoration, IPT \cite{chen2021pre} constructs a large-scale synthetic dataset on ImageNet for pre-training, equipped with multiple heads and multiple tails for multiple low-level vision tasks. Recently, Swin Transformer \cite{liu2021swin} presents a hierarchical Transformer structure with shifted window. The shifted window scheme decreases self-attention computational load and makes it possible to process high-resolution images. Uformer \cite{wang2021uformer} constructs a U-shaped Transformer network based on Swin Transformer for image restoration. However, limited by the computational cost of self-attention, existing Transformer-based structures are relatively shallow. Generally, for complex image restoration tasks, deeper network structure is superior in modeling more complicated noise patterns. Hence, inspired by U$^2$-Net \cite{qin2020u2}, we construct a two-level nested U-shaped Transformer structure, which enables the network to capture richer local and global context from both shallow and deep layers.

\noindent\textbf{Contrastive Learning.} Contrastive learning has been widely used in self-supervised representation learning \cite{doersch2015unsupervised, he2020momentum, chen2020simple}. Many previous works \cite{he2020momentum, chen2020simple, henaff2020data, grill2020bootstrap} have attempted to apply contrastive learning in high-level vision tasks, due to the inherent suitability for modeling feature contrasts between positive and negative samples. However, there are few works of applying contrastive learning to image restoration because of the difficulties in constructing contrastive samples and contrastive loss. Recently, \cite{wang2021unsupervised} introduce contrastive learning in blind Super-Resolution for learning abstract representations to distinguish various degradations in the representation space. \cite{wu2021contrastive} propose a novel contrastive regularization by utilizing both the information of hazy images and clear images as negative and positive samples, respectively. In this paper, We propose a novel multi-view contrastive learning scheme to further decouple the noise component from the background image.

\begin{figure}[!tp]
    \centering
    \begin{minipage}[b]{0.95\linewidth}
    \includegraphics[width=\linewidth]{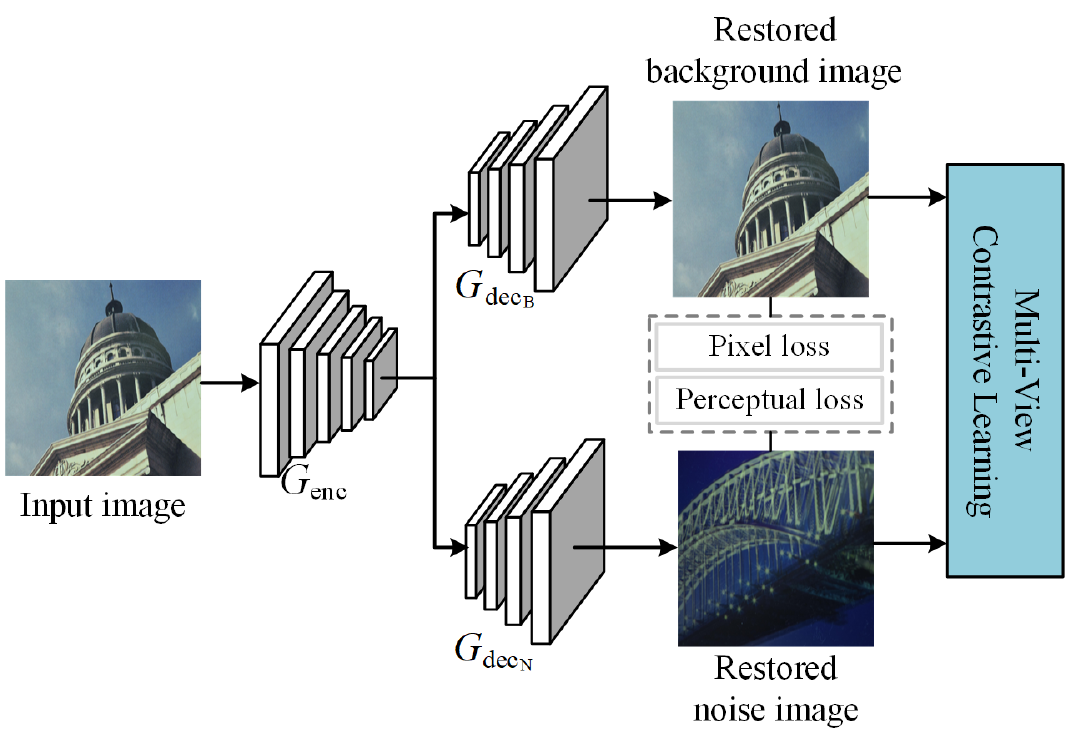}\vspace{4pt} 
    \end{minipage}
    \vspace{-10pt}
    \caption{Overall pipeline of the U$^2$-Former structure. The outer U-shaped Encoder-Decoder structure (Figure~\ref{fig2:RUT}(a)) consists of an Encoder $\mathbf{G}_\text{enc}$ and two parallel Decoders, including $\mathbf{G}_{\text{dec}_\text{B}}$ for decoupling the background features and $\mathbf{G}_{\text{dec}_\text{N}}$ for decoupling the noise features, respectively. Each stage in the U-shaped Encoder-Decoder contains a well-designed inner U-shaped Transformer block (Figure~\ref{fig2:RUT}(b)). A novel multi-view contrastive learning (Figure~\ref{fig_add:multi-view}) is utilized to guild the encoder to further decouple the noise component from the background image.
    }
    \label{fig1:pipline}
    \vspace{-8pt}
\end{figure}

\section{Method}

In this section, we will elaborate the proposed U$^2$-Former framework for image restoration. The overall pipeline of the U$^2$-Former is illustrated in Figure~\ref{fig1:pipline}. We first introduce the main architecture of the U$^2$-Former, and then describe how the contrastive learning guide U$^2$-Former to handle complex noise patterns. Finally, we describe specific supervision items for end-to-end parameter learning.

\begin{figure*}[!htbp]
    \centering
    \begin{minipage}[b]{0.95\linewidth}
    \includegraphics[width=\linewidth]{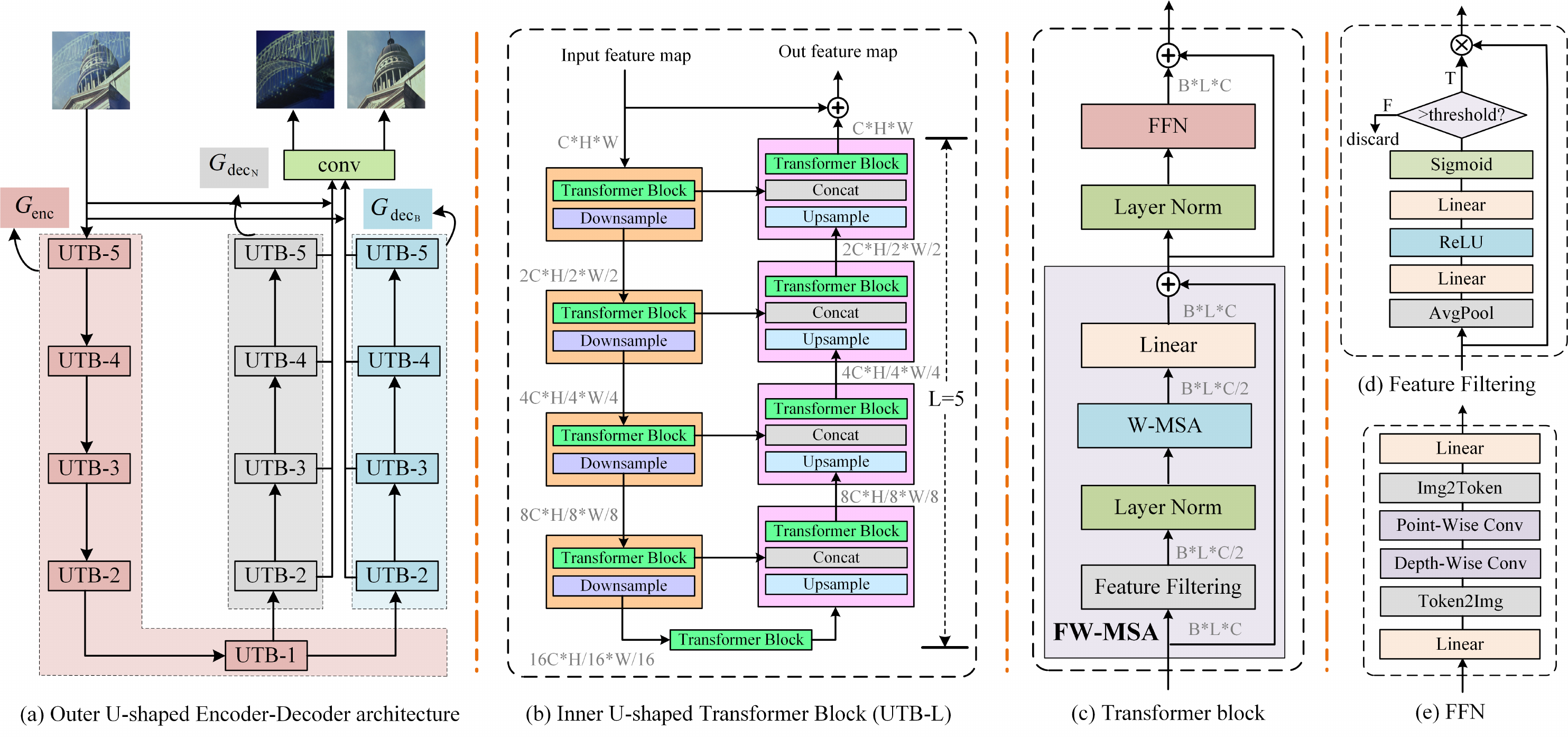}\vspace{4pt}
    \end{minipage}
    \vspace{-5pt}
    \caption{Main architecture our proposed U$^2$-Former for image restoration. (a) presents the outer U-shaped encoder-decoder structure, where each stage consists of an inner U-shaped Transformer Block (UTB) with different depth L as shown in (b). The detailed structure of the Transformer block in UTB is shown in (c), where we proposed a novel Feature-Filtering Window-based Multi-head Self-Attention (FW-MSA) for reducing the computational cost. The detailed structure of Feature-Filtering is illustrated in (d). Finally we introduce depthwise separable convolution in the Feed-Forward Network (FFN) for further capturing local dependencies as shown in (e).}
    \label{fig2:RUT}
    \vspace{-10pt}
\end{figure*}
\subsection{Network architecture of U$^2$-Former}
To overcome the limitations of existing Transformer-based structures in model depth and image reconstruction, in our U$^2$-Former, We embed multiple inner U-shaped Transformer blocks (UTB) inside the outer U-shaped encoder-decoder structure.

\subsubsection{Inner U-shaped Transformer block.}
The inner UTB is customized as the core operating unit in the U$^2$-Former, as shown in Figure~\ref{fig2:RUT}(b). 
Inspired by the CNN architecture in ~\cite{qin2020u2,szegedy2015going}, we design the UTB-L as a lightweight U-shaped Transformer architecture, where L denotes the depth of UTB. Specifically, the Transformer blocks (as shown in Figure~\ref{fig2:RUT}(c)), followed by downsampling and upsampling are introduced to achieve multi-scale feature maps from the input feature maps $\mathbf{X}\in \mathbb{R}^{C\times H\times W}$. Consequently, the global and local features of the image are further extracted, being beneficial for separating the noise component from the background branch. Besides, the mapped features $\mathbf{F(X)}$ from UTB is fused with the input feature maps $\mathbf{X}$ through a residual connection, so that more comprehensive information is $\mathbf{X+F(X)}$ is achieved.

\noindent\textbf{Feature Filtering Mechanism.}
To reduce computational cost of self-attention and avoid overfitting, we further propose a novel module named Feature-filtering Window-based Multi-head Self-Attention (FW-MSA) in Transformer block as shown in Figure~\ref{fig2:RUT}(c).
Specifically, it first utilizes the attention mechanism to obtain the attention weight of input features in the current feature dimension:
\begin{equation}
\begin{split}
    & \mathbf{W'}_i = \text{FC}(\text{AvgPool}(\mathbf{F}_i)), \\
    & \mathbf{W}_i = \sigma(\text{FC}(\delta(\mathbf{W'}_i))),
\end{split}
\end{equation}
where $\mathbf{F}_i$ is the feature belonging to the $i$-th feature dimension, $\mathbf{W}_i$ denotes the attention weight of the $i$-th feature dimension, $\sigma$ is the Sigmoid function and $\delta$ is the ReLU function.
Then, by empirically setting the threshold $\rho_{i}$ to filter the attention weights, our FW-MSA selects those features with high weights and distill more valuable information to compute self-attention:
\begin{equation}
    \mathbf{F}_s = 
        \begin{cases}
            \text{Concat}(\mathbf{F}_s, \mathbf{W}_{i}\!\cdot \mathbf{F}_{i}), & \mathbf{W}_{i} \textgreater\rho_{i} \\
            \mathbf{F}_{s} & \mathbf{W}_{i}\leq\rho_{i}
        \end{cases}
        ,
\end{equation}
where $\mathbf{F}_s$ denotes the selected features by our proposed feature-filtering. The detailed structure of the feature-filtering is shown in Figure~\ref{fig2:RUT}(d). To avoid overfitting, the residual connection with the input features $\mathbf{F}_i$ is added to the corresponding outputs:
\begin{equation}
\begin{split}
    & \mathbf{\hat{F}}_{s}=\text{W-MSA}\left(\text{LN}\left(\mathbf{F}_{s}\right)\right)+\mathbf{F}_{s}, \\
    & \mathbf{\hat{F}}_{i}=\text{FFN}\left(\text{LN}\left({\mathbf{\hat{F}}_{s}}\right)\right)+{\mathbf{F}_{i}},
\end{split}
\end{equation}
where the W-MSA denotes the window-based multi-head self-attention~\cite{liu2021swin}, LN is the layer normalization, and FFN is the feed-forward network as shown in Figure~\ref{fig2:RUT}(e). Importantly, the FFN here leverages depthwise separable convolution to capture local dependencies.

\subsubsection{Outer U-shaped Encoder-Decoder framework.} 
The outer U-shaped encoder-decoder framework is illustrated in Figure~\ref{fig2:RUT}(a). The encoder $G_{\text{enc}} $ consists of 5 stages, and two parallel decoders: $G_{\text{dec}_\text{B}}$ for decoupling background layer features and $G_{\text{dec}_\text{N}}$ for decoupling noise layer features, are both composed of 4 stages. Each stage contains a well-designed UTB-L. In the first four stages in $G_{\text{enc}}$, we employ UTB-5, UTB-4, UTB-3, and UTB-2, respectively. In fifth stage in $G_{\text{enc}}$, the resolution of the feature maps is already relatively low. To preserve the valuable features as much as possible, we remove the downsampling operation and merely stack 6 Transformer blocks. More specifically, given input feature maps $X_0\in \mathbb{R}^{C\times H\times W}$, the $l$-th stage of the encoder outputs the feature maps $X_l\in \mathbb{R}^{2^lC\times \frac {H}{2^l}\times \frac {W}{2^l}},l=\{1,2,3,4\}$. For the decoders, each stage has a symmetric structure compared with that in the encoder. We take the upsampled feature maps and ones from each symmetric encoder stage as input for the next stage in the decoder. Each decoder stage can generate a side feature maps. We then upsample these feature maps to ensure their size to be the same to the input image size and fuse them through concatenation. Finally we generate the reconstructed clear background image  and noise images by $1\times 1$ convolutional layer.
\subsection{Multi-view Contrastive Learning}
\begin{figure}[!tp]
    \centering
    \begin{minipage}[b]{0.95\linewidth}
    \includegraphics[width=\linewidth]{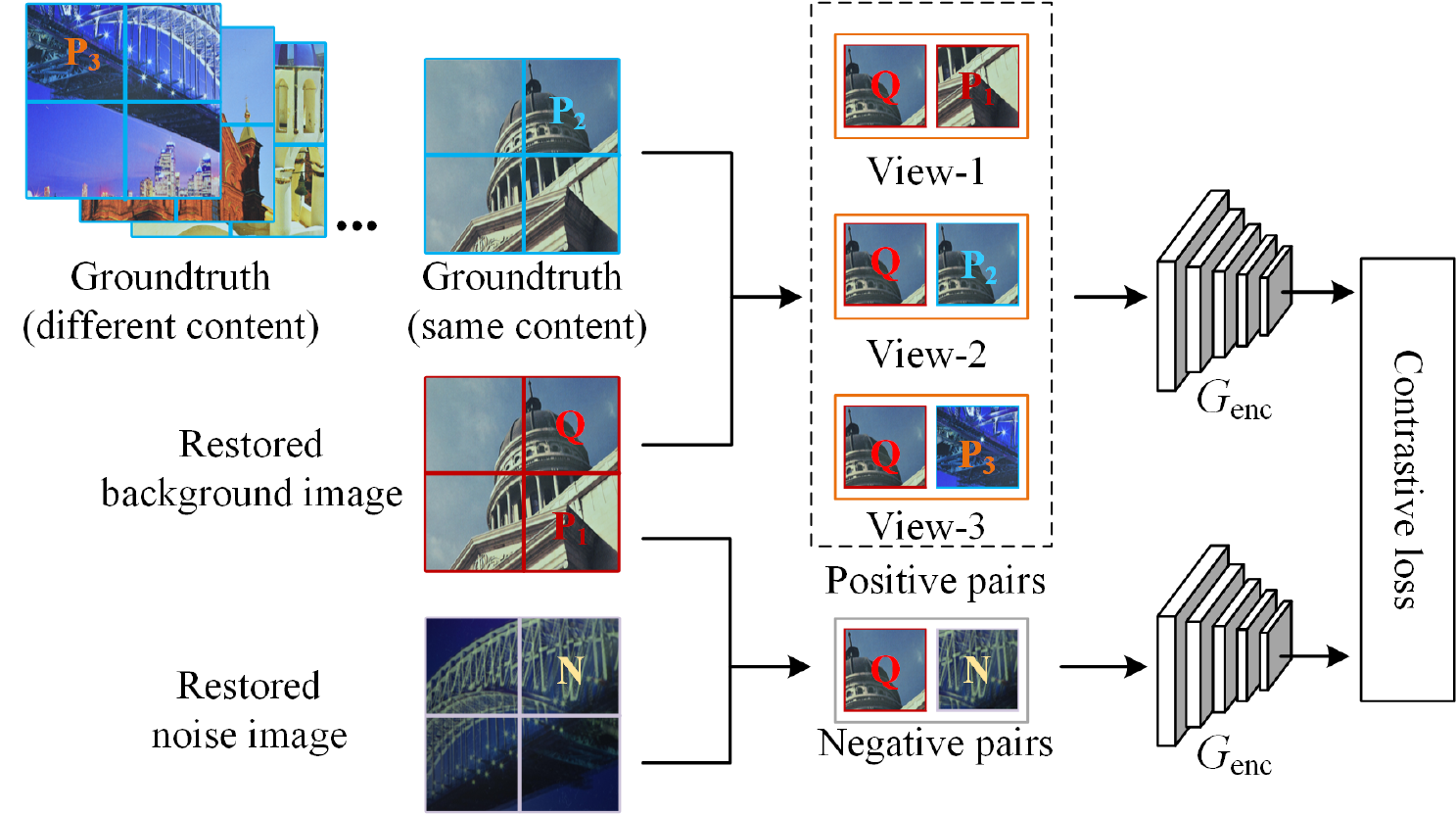}\vspace{4pt}
    \end{minipage}
    \caption{Illustration of our proposed multi-view contrastive learning. We construct three types of positive pairs (corresponding to three views of contrastive learning) including: \textbf{View}-1: patches from the same restored image; \textbf{View}-2: patches from the restored background image and the corresponding groundtruth background (with same image content) and \textbf{View}-3 patches from  the restored background image and a random groundtruth background image (with different image content). Besides, the negative pairs are constructed by comparing the restored background image and the restored noise image in all three cases.}
    \label{fig_add:multi-view}
    \vspace{-10pt}
\end{figure}

Contrastive learning is a discriminant-based approach that groups similar contents close and dissimilar contents away. Although it has demonstrated effectiveness in many high-level vision tasks, it still has great potential in image restoration due to the difficulties of semantic guidance in image synthesis. In the paper, we propose a novel multi-view contrastive learning scheme to guide our U$^2$-Former learning to remove complex noise patterns. As illustrated in Figure~\ref{fig_add:multi-view}, we first crop restored background image, noise image, and multiple groundtruth images from the same batch into patches. We then label background patches and groundtruth patches as positive samples, while noise patches as negative samples respectively. Contrastive learning are conducted in three aspects (views) when constructing positive pairs:
\begin{itemize}
    \item \textbf{View-1}: The noise distribution in the input image is not necessarily uniform. To ensure the restoring consistency between different regions in the same image, we take two patches from the same restored background image as positive pairs in view-1 contrastive learning.
    \item \textbf{View-2}: We pair the restored background image and the corresponding groundtruth background (with same image content) in patch level to be positive to guild the model to restore the clean background image.
    \item  \textbf{View-3}: We take the restored background image to a random groundtruth background image (with different image content) in patch level as positive pairs to encourage the model to focus on learning the features that are sensitive to noise rather than the image content.
\end{itemize}
The negative pairs are constructed in a fixed manner for all three views: comparing the restored background image and the restored noise image. In summary, through constructing contrastive pairs from multiple perspectives, we encourage our model to learn image degradation rather than similar image content. More specifically, we feed these patches to the encoder followed by extra two-layer fully connection named MLP to obtain feature embeddings for computing feature similarity:
\begin{equation}
    \mathbf{e}_{p_i} = \text{MLP}(G_{\text{enc}}(p_i)),
\end{equation}
where $\mathbf{e}_{p_i}$ denotes the feature embedding of $i$-th image patch $p_i$, and $G_{\text{enc}}$ is the encoder of the U$^2$-Former. In this way, the query patch, positive patches and negative patches are mapped into N-dimension vectors respectively. Thus, to maximize the mutual information between corresponding patches, we employ the noise contrastive estimation(NCE) framework~\cite{oord2018representation,park2020contrastive}, and then the binary classification is set up:
\begin{small}
\begin{equation}
    \mathcal{L}_{c}\!=\!-\!\sum_{i=1}^{N_i} \!\text{log}\!\left[\!\frac{\sum\limits_{j=1}^{N_j}\text{exp}(\mathbf{e}_{p_i}\!\cdot\!\mathbf{e}_{\text{pos}_j})}{\sum\limits_{j=1}^{N_j}\!\text{exp}(\mathbf{e}_{p_i}\!\cdot\!\mathbf{e}_{\text{pos}_j}\!)\!+\!\sum\limits_{k=1}^{N_k}\!\text{exp}(\mathbf{e}_{p_i}\!\cdot\!\mathbf{e}_{\text{neg}_k}\!)}\!\right]\!,\!
\end{equation}
\end{small}
where $N_i$, $N_j$, and $N_k$ are the number of patches for query, positive and negative sets respectively; $\mathbf{e}_{\text{pos}}$ and $\mathbf{e}_{\text{neg}}$ denote the feature embeddings of positive samples and negative samples.
\subsection{Jointly Supervised Parameter Learning}
We optimize the whole model of our U$^2$-Former in an end-to-end manner. 

\subsubsection{Multi-stage Pixel Reconstruction Loss.} We employ $L_1$ loss to push the pixel values of generated background image $\hat{\mathbf{T}}$ and noise image $\hat{\mathbf{R}}$ in various stages as close as their groundtruth. The multi-stage pixel reconstruction loss is formulated as follows:
\begin{small}
\begin{equation}
    \mathcal{L}_{\text{pixel}}=\alpha_1\sum_{i=1}^{N}{\theta_i\mathcal{L}_1\left(\mathbf{T},{\hat{\mathbf{T}}}_i\right)}+\beta_1\sum_{i=1}^{N}{\theta_i\mathcal{L}_1\left(\mathbf{R},{\hat{\mathbf{R}}}_i\right)},
\end{equation}
\end{small}
where $\mathbf{T}$ and $\mathbf{R}$ denote groundtruth of the background image and noise image respectively, ${\hat{\mathbf{T}}}_i$ and ${\hat{\mathbf{R}}}_i$ denote the reconstructed background image and noise image in the $i$-th stage, $\alpha_1$ and $\beta_1$ are the weights of each loss term, $\theta_i$ is the weight of each $L_1$ loss term, and $N$ is the number of stages of the decoder. Empirically, $\theta=\{ \theta_i |0.1,0.1,0.1,0.7, i=1,2,3,4\}$, $\alpha_1=0.7$ and $\beta_1=0.3$.

\subsubsection{Multi-stage Perceptual Loss.} The perceptual loss~\cite{johnson2016perceptual} is proposed to perform semantic supervision on generated images in the deep feature space. Specifically, the pretrained VGG-19~\cite{simonyan2014very} is exploited as the feature extractor:
\begin{small}
\begin{equation}
    \mathcal{L}_\text{p}=\alpha_2\sum_{i=1}^{N}{\theta'_i\mathcal{L}_{\text{VGG}}\left(\mathbf{T},{\hat{\mathbf{T}}}_i\right)}+\beta_2\sum_{i=1}^{N}{\theta'_i\mathcal{L}_{\text{VGG}}\left(\mathbf{R},{\hat{\mathbf{R}}}_i\right)},
\end{equation}
\end{small}
where $\mathcal{L}_{\text{VGG}}$ denotes perceptual distance between two deep features. Empirically, $\theta'=\{ \theta'_i |0.1,0.1,0.1,0.7, i=1,2,3,4\}$, $\alpha_2=0.7$ and $\beta_2=0.3$.

Therefore, the overall training loss is defined as:
\begin{equation}
    \mathcal{L}=\lambda_1\mathcal{L}_{\text{pixel}}+\lambda_2\mathcal{L}_{\text{p}}+\lambda_3{L}_{\text{c}},
\end{equation}
where $\lambda_1, \lambda_2,$ and $\lambda_3$ are the hyper-parameters to balance various losses, and we empirically set them as $\lambda_1=1,\lambda_2=0.2,\lambda_3=0.5$ respectively.

\section{Experiments}
To evaluate the performance and generalizability of our proposed U$^2$-Former, we carry out experiments on multiple challenging image restoration tasks, including (1) image reflection removal, (2) image deraining, and (3) image dehazing. Besides we perform indispensable ablation study on the image reflection removal task to analyze the contribution of the key components in our model.
\begin{figure}[!bp]
    \centering
    \vspace{-5pt}
    \begin{minipage}[b]{0.7\linewidth}
    \includegraphics[width=\linewidth]{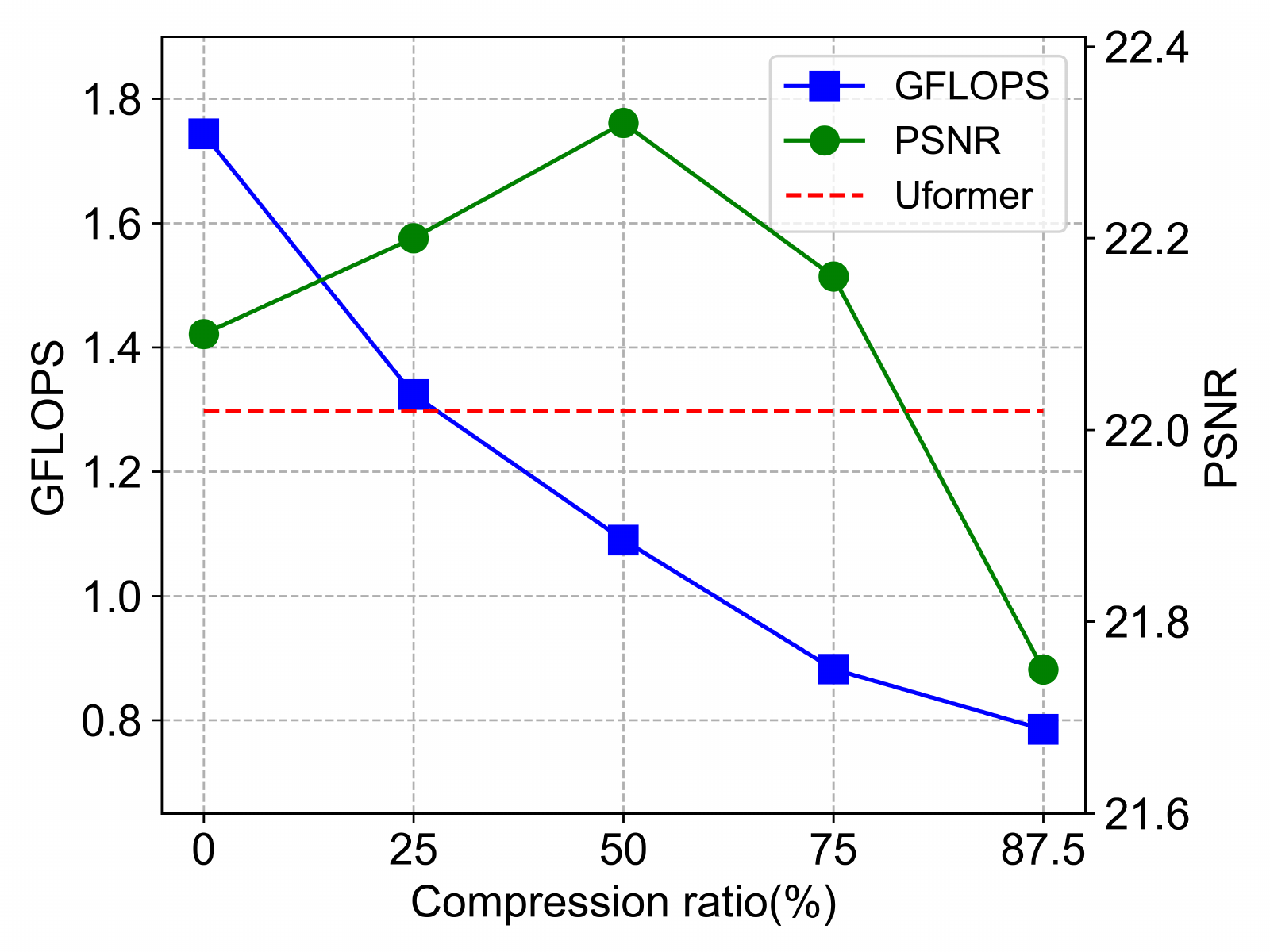}\vspace{4pt}
    \end{minipage}
    \vspace{-10pt}
    \caption{Effect of varying compression ratio on Uformer-UTB by our Feature Filtering Mechanism. Here the computational cost and the compression ratio are both calculated for one Transformer block in Uformer-UTB. The performance of Uformer is provided for reference.}
    \label{fig_add2:compress}
    \vspace{-8pt}
\end{figure}
\begin{figure*}[!t]
    \centering
    \begin{minipage}[b]{0.95\linewidth}
    \includegraphics[width=\linewidth]{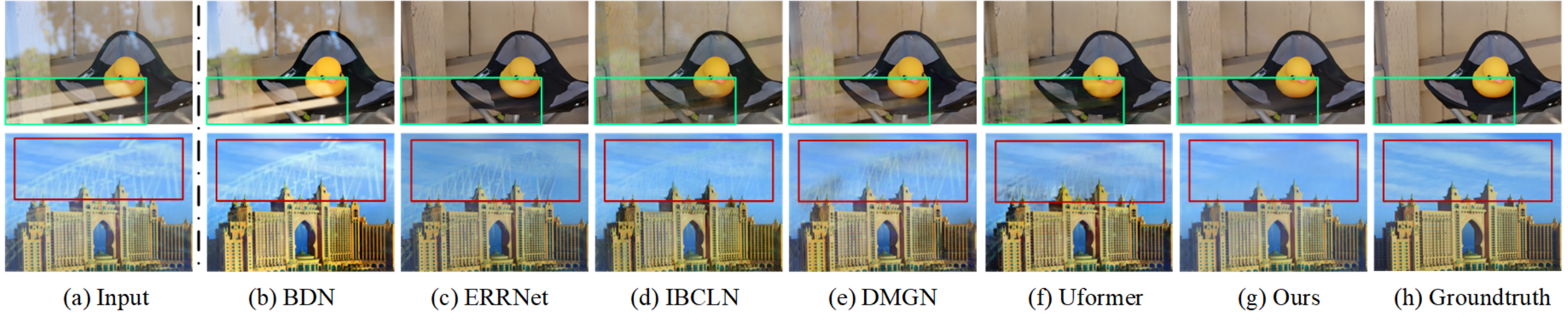}\vspace{4pt}
    \end{minipage}
    \vspace{-8pt}
    \caption{Qualitative comparison of reflection removal between different methods. Our model remove most of the undesired reflections and retain more high frequency details than other methods, particularly in the regions indicated by bounding boxes.}
    \label{fig5:dereflection}
    \vspace{-12pt}
\end{figure*}
\subsection{Experimental setup}
\textbf{Evaluation metrics.} We employ the commonly used PSNR and SSIM metrics to measure the performance of our proposed model. The higher value of PSNR or SSIM denotes the better restoration result.

\noindent\textbf{Implementation Details.} We train different models for three different tasks with the following settings. We implement our model in distribution mode with 6 GeForce RTX 3090 under the pytorch framework. The Adam optimizer~\cite{kingma2014adam} with the initial learning rate $2\times10^{-4}$ is employed for optimization. The Transformer block window is set to $8\times8$ and the maximum of epochs is set to 250. In particular, all images are resized to 512$\times$512 for test. Like most existing methods, random flipping, random cropping and resizing are used for data augmentation.

\subsection{Dataset}
\noindent\textbf{Image Reflection Removal.} We generate 5,754 synthetic image pairs using the method proposed by PLNet~\cite{zhang2018single}. Besides, 90 real image pairs from PLNet, and 200 real image pairs from IBCLN~\cite{li2020single} are also used for training. Five real-world datasets including Real20, Nature, Solid, Wild, and Postcard are employed for testing.

\noindent\textbf{Image Deraining.} We perform experiments on two well-known datasets for rain streak removal, including Rain100H and Rain100L~\cite{yang2017deep}. Rain100H consists of heavy rainy images, in which 1,800 images are for training and 100 images are for testing. Rain100L contains images under the light rainy case, where 1,800 and 100 images are split for training and testing respectively. 

\noindent\textbf{Image Dehazing.} We conduct experiments on a popular benchmark dataset named RESIDE~\cite{li2018benchmarking} for image dehazing. For testing, the Synthetic Objective Testing Set (SOTS) in RESIDE, which contains 500 indoor and 500 outdoor images, is regarded as the testing set.

\subsection{Ablation Study}
In this section, we conduct experiments on the reflection removal task to investigate the proposed functional techniques in our U$^2$-Former. Therefore, we perform ablation study on the Real20 dataset to compare our U$^2$-Former to its five related variants:
1) \textbf{U-Net}~\cite{ronneberger2015u}, which is designed based on CNNs and adopts one single U-shaped structure; 2) \textbf{U$^2$-Net}~\cite{qin2020u2} that is also based on CNN but employs two nested U-shaped structures; 
3) \textbf{Uformer}~\cite{wang2021uformer} that is designed in one single u-shaped structure based on Transformer; 4) \textbf{Uformer-UTB}, which replaces the Transformer block of Uformer with our proposed inner U-shaped Transformer Block (UTB); 5) \textbf{U$^2$-Former(w/o CL)}, i.e., our U$^2$-Former without contrastive learning.

Table~\ref{tab2:ablation} shows the experimental results of the ablation study. Comparing Uformer to Uformer-UTB, the result demonstrates that our proposed inner U-shaped Transformer Block (UTB) leads to better performance than the Transformer block in Uformer with less computation cost. Further, when we stack more Transformer blocks, the performance of the U$^2$-Former(w/o CL) gains an obvious improvement. Finally, U$^2$-Former (ours) achieves satisfactory enhancement compared to U$^2$-Former (w/o CL), proving that our proposed multi-view contrastive learning is effective in guiding the model to decouple the noise component from the background image. An interesting observation is that the comparisons between the CNN-based methods (U-Net and U$^2$-Net) and the Transformer-based methods (Uformer and U$^2$former) imply the advantages of Transformer over CNNs for feature learning on image restoration.
\begin{table}[!htbp]
    \centering
    \caption{Ablation study on our U$^2$-Former in terms of PSNR and SSIM to investigate the effectiveness of each proposed technique in our model.}
    \renewcommand\arraystretch{1.0}
    \vspace{-5pt}
    \resizebox{1.0\linewidth}{!}{
    \begin{tabular}{l|cc|cc}
    \toprule
    Method  &\makecell[c]{Transformer\\ block\\} &\makecell[c]{Contrastive\\ learning\\} & PSNR  & SSIM \\
    \midrule
U-Net 	&\XSolidBrush	&\XSolidBrush	&19.85 & 0.763 \\ 
U$^2$-Net 	&\XSolidBrush	&\XSolidBrush	&20.13 &0.771 \\
Uformer     &\CheckmarkBold	&\XSolidBrush     &22.02 &0.776 \\
Uformer-UTB 	&\CheckmarkBold	&\XSolidBrush	&22.32 &0.789 \\
U$^2$-Former (w/o CL)	&\CheckmarkBold	&\XSolidBrush	&22.96 &0.818 \\
U$^2$-Former (ours) &\CheckmarkBold	&\CheckmarkBold	&\textbf{23.67} &\textbf{0.835} \\
\bottomrule
    \end{tabular}
    }
    \label{tab2:ablation}
\end{table}


\noindent\textbf{Effect of varying the compression ratio by the feature-filtering mechanism.} As shown in Figure~\ref{fig_add2:compress}, we investigate the effect of compression ratio on computational cost and model performance in Uformer-UTB. Here we only statistically analyze the computational cost of one Transformer Block. When our proposed feature-filtering percolates 50$\%$ features, the model performance is improved instead and the computational cost is greatly reduced. It sufficiently demonstrates that our proposed feature-filtering mechanism can reduce the computational requirements while distilling more valuable features for self-attention.

\begin{figure*}[!tp]
    \centering
    \begin{minipage}[b]{0.95\linewidth}
    \includegraphics[width=\linewidth]{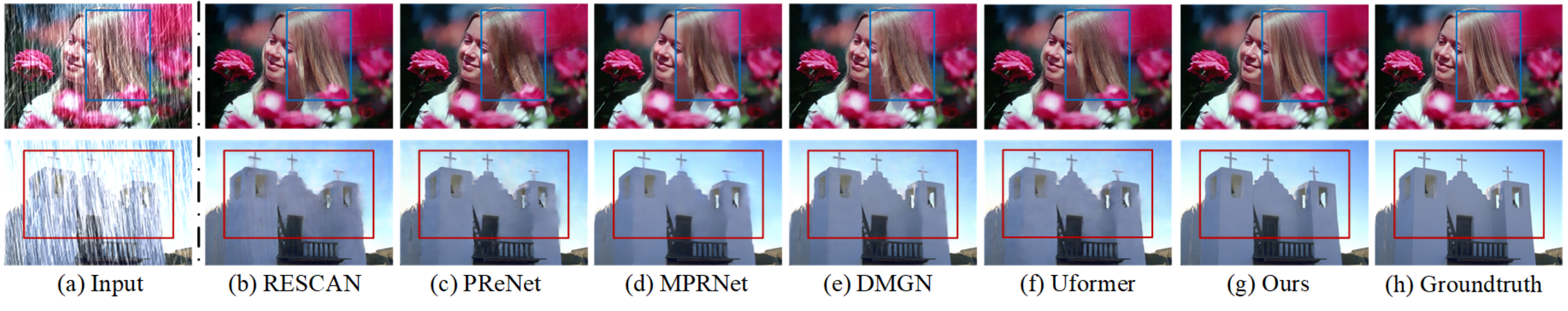}
    \end{minipage}
    \caption{Visual comparison on the Rain100 dataset for rain streak removal. Our U$^2$-Former can recover more details in the reconstructed images. Best viewed in zoom-in mode.}
    \label{fig6:deraining}
\end{figure*}

\begin{figure*}[!t]
    \centering
    \begin{minipage}[b]{0.95\linewidth}
    \includegraphics[width=\linewidth]{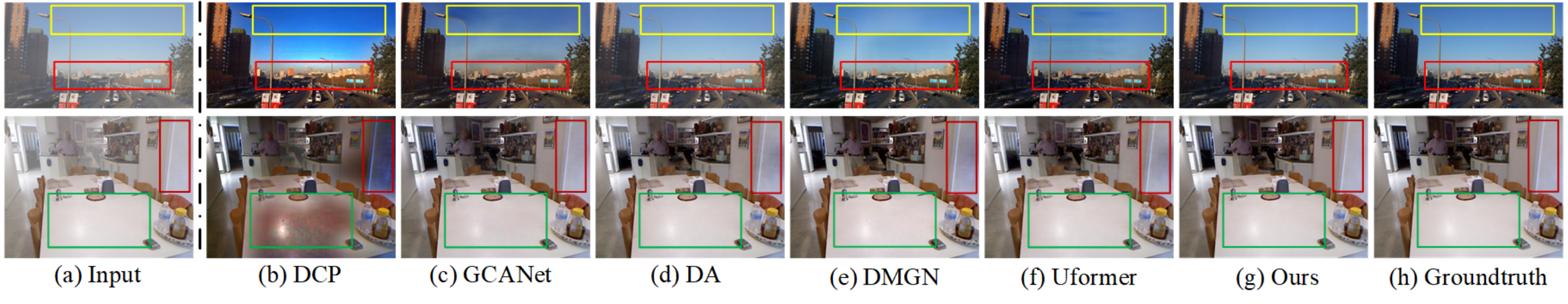}
    \end{minipage}
    \caption{Dehazing results on the SOTS dataset. Compared to the sate-of-the-art methods, our U$^2$-Former effectively mitigate the color distortion and generates images are visually closer to the groundtruth. Best viewed in zoom-in mode.}
    \label{fig7:dehaze}
    \vspace{-5pt}
\end{figure*}

\subsection{Experiments on Image Reflection Removal.}
Table~\ref{tab3:reflection} presents the quantitative results of different models for reflection removal on five real-world datasets. We conduct experiments to compare our model with 8 sate-of-the-art methods: YW19~\cite{yang2019fast}, BDN~\cite{yang2018seeing}, RmNet~\cite{wen2019single}, ERRNet~\cite{wei2019single}, Kim~\cite{kim2020single}, IBCLN~\cite{li2020single}, DMGN~\cite{feng2021deep}, and Uformer~\cite{wang2021uformer}. It is true that IBCLN is slightly superior to our U$^2$-Former on the Postcard dataset. However, our method outperforms other methods on all other datasets. The visualization results are shown in Figure~\ref{fig5:dereflection}. Here we randomly select several cases with strong reflections from different scenes. Compared with other methods, our model can remove most of the undesired reflections and retain more high frequency details.
\begin{table}[!htbp]
    \centering
    \caption{Quantitative results of different models for image reflection removal in 5 real-world datasets in terms of PSNR and SSIM.}
    \vspace{-5pt}
    \renewcommand\arraystretch{1.0}
    \resizebox{1.0\linewidth}{!}{
    \Large
    \begin{tabular}{l|ccccccccccc}
    \toprule
    \multirow{2}{*}{Method} & \multicolumn{2}{c}{Real20} & \multicolumn{2}{c}{Nature} & \multicolumn{2}{c}{Solid} & \multicolumn{2}{c}{Wild} & \multicolumn{2}{c}{Postcard} \\
    \cmidrule(lr){2-3}
    \cmidrule(lr){4-5}
    \cmidrule(lr){6-7}
    \cmidrule(lr){8-9}
    \cmidrule(lr){10-11}
    & PSNR & SSIM & PSNR & SSIM & PSNR & SSIM & PSNR & SSIM & PSNR & SSIM \\
    \midrule
YW19 &16.80 &0.547 &16.60 &0.608 &16.65 &0.646 &18.54 &0.702 &17.14 &0.674 \\ 
BDN &22.18    &0.816    &20.74    &0.801    &19.84    &0.831    &20.02    &0.827    &18.71    &0.772    \\
RmNet &18.54    &0.707    &19.07    &0.755    &20.74    &0.820    &22.02    &0.833    &20.08    &0.831    \\
ERRNet &23.41    &0.832    &20.79    &0.796    &24.86    &0.903    &23.45    &0.833    &21.49    &0.870    \\
Kim & 19.41    & 0.715    & 20.56    & 0.787    & 23.86    & 0.889    & 24.94    & 0.889    & 22.80    &0.874    \\
IBCLN &22.03    &0.789    &23.71    &0.820    &24.99    &0.901    &23.98    &0.886    &\textbf{23.24}    &0.877    \\
DMGN &22.11 &0.805 &19.63 &0.812 &24.43 &0.891 &24.72 & 0.887 &22.91 &0.884 \\
Uformer &22.02 &0.776 &23.07 &0.823 &24.52  &0.897 & 23.96 & 0.878 & 21.08 & 0.849 \\
\textbf{Ours} &\textbf{23.67}    &\textbf{0.835}   &\textbf{24.75}    &\textbf{0.848}    &\textbf{25.27}    &\textbf{0.907}    &\textbf{25.68}    &\textbf{0.905}    &22.43    &\textbf{0.889}    \\
\bottomrule
    \end{tabular}
    }
    \label{tab3:reflection}
    \vspace{-5pt}
\end{table}

\subsection{Experiments on Image Deraining.}
We conduct deraining experiments on two synthetic datasets to evaluate the performance of our proposed model. Table~\ref{tab4:derain} presents the quantitative results of our model compared with other the-state-of-art methods, including GMM~\cite{li2016rain}, DDN~\cite{fu2017removing}, DID-MDN~\cite{zhang2018density}, RESCAN~\cite{li2018recurrent}, PReNet~\cite{ren2019progressive}, MPRNet~\cite{zamir2021multi}, IPT~\cite{chen2021pre}, DMGN~\cite{feng2021deep}, and Uformer~\cite{wang2021uformer}. It can be clearly seen that our method achieves the best performance (39.31dB) with the 1.63dB improvement on rain100L, and the best results (30.87dB) with the 1.50dB improvement on rain100H, respectively.
Figure~\ref{fig6:deraining} shows the visualization results. Obviously, existing methods suffer from over-smoothing and blurring in image details (marked by the bounding box in Figure~\ref{fig6:deraining}). However, our model can recover more details in the reconstructed images.
\begin{table}[!htbp]
    \vspace{-5pt} 
    \centering
    \caption{Quantitative results of different models for rain streak removal on two datasets in terms of PSNR and SSIM. }
    \renewcommand\arraystretch{1.0}
    \resizebox{0.7\linewidth}{!}{
    \begin{tabular}{l|cccc}
    \toprule
    \multirow{2}{*}{Method} & \multicolumn{2}{c}{Rain100L} & \multicolumn{2}{c}{Rain100H} \\
    \cmidrule(lr){2-3}
    \cmidrule(lr){4-5}
    & PSNR & SSIM & PSNR & SSIM \\
    \midrule
GMM &25.66	&0.733	&14.38	&0.434 \\ 
DDN &28.80	&0.905	&16.80	&0.541 \\
DID-MDN &24.30	&0.823	&15.99	&0.551 \\
RESCAN &37.23	&0.978	&28.05	&0.868 \\
PRENet &36.83	&0.976	&28.73	&0.880 \\
MPRNet	&33.72	&0.950	&28.32	&0.863 \\
IPT    &35.10  &0.980   &- &- \\
DMGN   &36.24 &0.972     & 27.67 & 0.856 \\
Uformer &37.68 &0.976 &29.37 &0.877 \\
\textbf{Ours} &\textbf{39.31}    &\textbf{0.982}   &\textbf{30.87}  &\textbf{0.899} \\
\bottomrule
    \end{tabular}
    }
    \label{tab4:derain}
    \vspace{-12pt}
\end{table}

\subsection{Experiments on Image Dehazing.}
We conduct dehazing experiments to compare our U$^2$-Former with other sate-of-the-art methods: DCP~\cite{he2010single}, DahazeNet~\cite{cai2016dehazenet}, AOD-Net~\cite{li2017aod}, GCANet~\cite{chen2019gated}, DA~\cite{shao2020domain}, DM$^2$F-Net~\cite{deng2019deep}, DMGN~\cite{feng2021deep}, and Uformer~\cite{wang2021uformer}. The quantitative results on the RESIDE dataset are listed in Table~\ref{tab:dehaze}. Our model reaches the best performance for both indoor and outdoor scenes. Especially on the outdoor dataset, it gains an improvement of 2.54dB. 
\begin{table}[!htbp]
\vspace{2pt}
    \centering
    \caption{Quantitative results of different models for image dehazing in synthetic datasets in terms of PSNR and SSIM.}
    \renewcommand\arraystretch{1.0}
    \resizebox{0.7\linewidth}{!}{
    \begin{tabular}{l|cccc}
    \toprule
    \multirow{2}{*}{Method} & \multicolumn{2}{c}{Indoor} & \multicolumn{2}{c}{Outdoor} \\
    \cmidrule(lr){2-3}
    \cmidrule(lr){4-5}
    & PSNR & SSIM & PSNR & SSIM \\
    \midrule
DCP &19.91	&0.857	&17.12	&0.844 \\ 
DehazeNet &21.45	&0.864	&22.90	&0.881 \\
AOD-Net &17.68	&0.785	&19.62	&0.836 \\
GCANet &29.72	&0.953	&23.18	&0.914 \\
DA	&25.66	&0.927	&27.14	&0.938 \\
DM$^2$F-Net &34.27 &0.965 &26.50 &0.912 \\
DMGN & 30.95 &0.977 &28.56 &0.970 \\
Uformer &31.91 &0.971 &26.52 &0.944 \\
\textbf{Ours} &\textbf{36.42}    &\textbf{0.988}   &\textbf{31.10}  &\textbf{0.976} \\
\bottomrule
    \end{tabular}
    }
    \label{tab:dehaze}
    \vspace{-10pt}
\end{table}
The visualization results are shown in Figure~\ref{fig7:dehaze}. The dehazing images obtained by our method evidently mitigate the color distortion compared with previous methods (such as DCP, DA), obtaining much better visualization in some details. More experimental results and user studies are available in the supplemental material.

\section{Conclusion}
In this work, we propose a general Transformer architecture, termed as U$^2$-Former, for image restoration. Our U$^2$-Former adopts two nested U-shaped structure to facilitate the information communication across different layers, Besides, we propose the Feature Filtering Mechanism to optimize the computational efficiency for the basic Transformer block. Consequently, our model is able to construct deep encoding and decoding space based on Transformer block. To further decouple the noise component and the background component, our U$^2$-Former performs multi-view contrastive learning by constructing the positive pairs in multiple aspects. 
Experimental results on three image restoration tasks demonstrate our superiority against the state-of-the-art methods for image restoration. 

\clearpage
\bibliography{main}
\end{document}